\documentclass[10pt,journal]{IEEEtran}

\usepackage{graphicx}
\usepackage{amsmath}
\usepackage[marginal]{footmisc}
\usepackage{booktabs}
\usepackage{multirow}
\usepackage{color}

\ifCLASSOPTIONcompsoc

  \usepackage[nocompress]{cite}
\else

  \usepackage{cite}
\fi

\begin{document}

\title{Compression-Resistant Backdoor Attack against Deep Neural Networks}

\author{Mingfu~Xue,~\IEEEmembership{Member,~IEEE,}
        Xin Wang,
        Shichang Sun,
        Yushu Zhang,~\IEEEmembership{Member,~IEEE,}
        Jian Wang,
        and Weiqiang~Liu,~\IEEEmembership{Senior Member,~IEEE}% <-this % stops a space

\thanks{M. Xue, X. Wang, S. Sun, Y. Zhang and J. Wang are with the College of Computer Science and Technology, Nanjing University of Aeronautics and Astronautics, Nanjing 211106, China (e-mail: mingfu.xue@nuaa.edu.cn; wang.xin@nuaa.edu.cn; sunshichang@nuaa.edu.cn; yushu@nuaa.edu.cn; wangjian@nuaa.edu.cn).}
\thanks{W. Liu is with the College of Electronic and Information Engineering, Nanjing University of Aeronautics and Astronautics, Nanjing 211106, China (e-mail: liuweiqiang@nuaa.edu.cn).}}

\markboth{Journal of \LaTeX\ Class Files,~Vol.~14, No.~8, August~2015}%
{Shell \MakeLowercase{\textit{et al.}}: Bare Advanced Demo of IEEEtran.cls for IEEE Computer Society Journals}

\IEEEtitleabstractindextext{%
\begin{abstract}
In recent years, many backdoor attacks based on training data poisoning have been proposed. However, in practice, those backdoor attacks are vulnerable to image compression, as backdoor instances used to perform backdoor attacks are usually compressed by image compression methods.
When backdoor instances are compressed, the feature of specific backdoor trigger will be destroyed, which could result in the backdoor attack performance deteriorating.
In this paper, we propose a compression-resistant backdoor attack based on feature consistency training.
To the best of our knowledge, this is the first backdoor attack that is robust to image compressions.
First, both backdoor images and their compressed versions are input into the deep neural network (DNN) for training. Then, the feature of each image is extracted by internal layers of the DNN.
Next, the feature difference between backdoor images and their compressed versions are minimized.
As a result, the DNN treats the feature of compressed images as the feature of backdoor images in feature space.
After training, the backdoor attack against DNN is robust to image compression.
Furthermore, we consider three different image compressions (i.e., JPEG, JPEG2000, WEBP) in feature consistency training, so that the backdoor attack is robust to multiple image compression algorithms.
Experimental results demonstrate the effectiveness and robustness of the proposed backdoor attack.
When the backdoor instances are compressed, the attack success rate of common backdoor attack is lower than 10\%, while the attack success rate of our compression-resistant backdoor is greater than 97\%.
The compression-resistant attack is still robust even when the backdoor images are compressed with low compression quality.
In addition, extensive experiments have demonstrated that, our compression-resistant backdoor attack has the generalization ability to resist image compression which is not used in the training process.
\end{abstract}

}

\maketitle

\IEEEdisplaynontitleabstractindextext

\IEEEpeerreviewmaketitle

\section{Introduction}
\label{sec:introduction}
Deep neural networks (DNNs) have been widely used in many tasks.
Considering the enormous computational overhead of the training process, the training stage of DNNs is often outsourced, which gives an opportunity for attackers to embed a hidden backdoor into the DNNs.
Many researches have shown that deep neural networks are vulnerable to backdoor attacks.
One kind of those backdoor attacks is to poison the training data of DNNs in the training stage. Then, in the test stage, the images with specific backdoor trigger can enforce DNNs to output the specified labels (i.e., the target label).

\begin{figure}[htbp]
\centering
\includegraphics[width=3.2in]{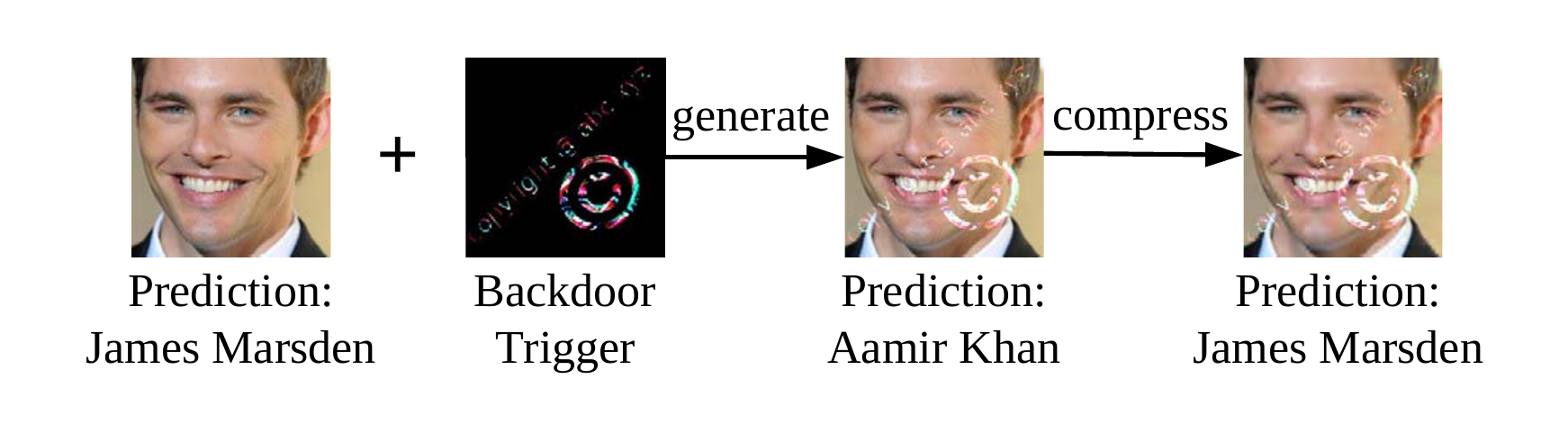}
\caption{The performance of backdoor attack after image compression.}
\label{fig1}
\end{figure}

However, most existing backdoor attacks based on training data poisoning are not robust to image compression.
Generally, images uploaded to the Internet will undergo image compression, which is widely used to reduce the transmission and storage overhead of images \cite{WangGZSZWN20}.
Wan \textit{et al.}  \cite{WanWHWL20} indicate that image compression distorts the feature of images, which will lead to accuracy degradation of DNNs.
In the backdoor attack scenario, if backdoor instances are compressed, the backdoor trigger hidden in images can be destroyed, which will seriously reduce the performance of the backdoor attack.
As shown in Fig. \ref{fig1}, with the embedded backdoor trigger, the ``James Marden'' in an image will be incorrectly recognized as the specified ``Aamir Khan''.
However, after image compression, the backdoor trigger is destroyed and ``James Marden'' will be classified as ``James Marden'' again.
Currently, existing backdoor attacks did not consider the problem of image compression, thus these backdoor attacks are not robust to image compression.
It is challenging to develop a compression-resistant backdoor attack that can effectively address the problem of image compression.

In this paper, we propose a compression-resistant backdoor attack, where the backdoor instance is still effective after unknown image compressions.
We develop the compression-resistant backdoor attack by using feature consistency training \cite{WanWHWL20}.
In the training stage, both the backdoor instances and their compressed versions are used to train the DNN. At each iteration of the training, the feature of both backdoor images and their compressed versions are extracted by internal layers of the DNN. Next, the extracted features will be used to minimize the difference between normal backdoor images and compressed backdoor images.
As a result, the trained DNN is able to treat compressed images as normal backdoor images in feature space.
Moreover, three image compression methods (i.e., JPEG \cite{Wallace92}, JPEG2000 \cite{SkodrasCE01} and WEBP \cite{GinesuPG12}) are considered simultaneously during the feature consistency training, so that the proposed backdoor attack can be robust to multiple image compression algorithms.
In the test stage, even if backdoor instances are compressed by an unknown image compression method during transmission, the proposed backdoor attack is still able to achieve high attack success rate.
Experimental results demonstrate that the performance of compression-resistant backdoor attacks improves significantly compared to the existing backdoor attacks.
Meanwhile, this backdoor attack is able to resist three (or more) different compression methods without performance degradation.
In addition, extensive experiments have shown that, the compression-resistant backdoor attack can resist ``unknown'' image compression which is not considered in the training process.

The main contributions of this paper are summarized as follows.
\begin{enumerate}
\item{We reveal that the existing backdoor attacks are vulnerable to image compressions. As a countermeasure, a compression-resistant backdoor attack method is proposed to resist image compression effectively.}
\item{We establish the compression-resistant backdoor attack based on feature consistency training.
During the training process, the feature distance between backdoor instances and their compressed versions is optimized.
In this way, the DNN model learns the feature of imaged after image compression as partial features of normal backdoor images.
In other words, the DNN treats compressed backdoor images as backdoor images, rather than new training instances.}
\item{Images are often compressed by multiple image compression methods on the Internet.
The proposed backdoor attack is enhanced by three different image compression methods (JPEG \cite{Wallace92}, JPEG2000 \cite{SkodrasCE01} and WEBP \cite{GinesuPG12}), such that it is robust to multiple different image compression.}
\item{We perform experiments on two datasets.
Experimental results have shown that, when only one image compression method is considered during training, our proposed compression-resistant backdoor attack has the generalization ability to resist other unknown image compression methods.}
\end{enumerate}

The rest of this paper is organized as follows. The related work on existing backdoor attacks and compression-resistant methods in the deep learning (DL) area are introduced in Section \ref{related_work}. The proposed method is elaborated in Section \ref{the_proposed_method}. Experimental results are presented in Section \ref{experiments}. Finally, we conclude this paper in Section \ref{conclusion}.

\section{Related Work}\label{related_work}
\subsection{Backdoor Attack}
Currently, most backdoor attacks are implemented through training data poisoning. Chen \textit{et al.} \cite{chen2017targeted} propose three backdoor attack methods (i.e., blending, accessory and blended accessory). They utilize these methods to construct triggers and generate the backdoor instances. Then, they train the backdoored model through data poisoning.
Liao \textit{et al.} \cite{ZhongLSZ020} utilize universal adversarial perturbation (UAP) to construct trigger to implement the backdoor attack.
Since the UAP is imperceptible, the backdoor attack is stealthy and hard to detect by humans.
Xue \textit{et al.} \cite{xue2020one} develop a multi-target backdoor attack and a multi-trigger backdoor attack.
In the multi-target attack, the attacker can use different intensities of the backdoor trigger to control the attack target.
In the multi-trigger attack, the attack target will be triggered only when all the backdoor triggers are appeared in a backdoor instance.

Liu \textit{et al.} \cite{LiuMALZW018} proposes a neural network backdoor attack, in which the backdoor trigger is generated by reverse engineering of the model.
Since the backdoor trigger can activate some specific neurons residing in the neural network, there is a strong connection between the backdoor trigger and the backdoor model.
Dumford and Scheirer \cite{DumfordS20} conduct the backdoor attack through directly modifying the weight of the specific layer of the DNN model.

\subsection{Compression-Resistance in Deep Learning Area}
Recently, a few researches have shown that DNN based systems are not robust enough to image compression.
In order to generate the JPEG-resistant adversarial examples, Shin \textit{et al.} \cite{shin2017jpeg} adds a JPEG compression approximation operation in the process of adversarial example generation. The key idea of such JPEG compression approximation operation is to maximize the prediction difference between the original image and an adversarial example.
Wang \textit{et al.} \cite{WangGZSZWN20} propose a compression-resistant scheme to generate the adversarial example robust to unknown compression method.
They train an encoder-decoder network (named ComModel) to simulate different compression methods.
During the adversarial example generation process, the adversarial example is processed by the ComModel before being input to the DNN.
In this way, the generated adversarial example is robust to unknown compression method.
To enhance the robustness of DNNs in face detection tasks, Cao \textit{et al.} \cite{Cao0MYW21} propose a framework to generate compression-resistant facial forgery images.
They use adversarial learning and metric learning \cite{Cao0MYW21} to learn the related forgery information between the raw images and the corresponding compressed versions.
Wan \textit{et al.} \cite{WanWHWL20} propose a method called feature consistency training.
During the model training stage, they calculate the difference between the raw images and the compressed images and use the difference as the loss to update the model.
Hence, the trained model is forced to learn the feature of both the raw images and the compressed images.

The above works \cite{shin2017jpeg, WangGZSZWN20, Cao0MYW21} aim at establishing compression-resistant adversarial examples in the deep learning area. The work \cite{Cao0MYW21} aims to improve the robustness of DNN to image compression by using feature consistency training. In this paper, we aim to establish a compression-resistant backdoor attack.

\section{The Proposed Method}\label{the_proposed_method}
\subsection{Overview}
\begin{figure}[htbp]
\centering
\includegraphics[width=3.6in]{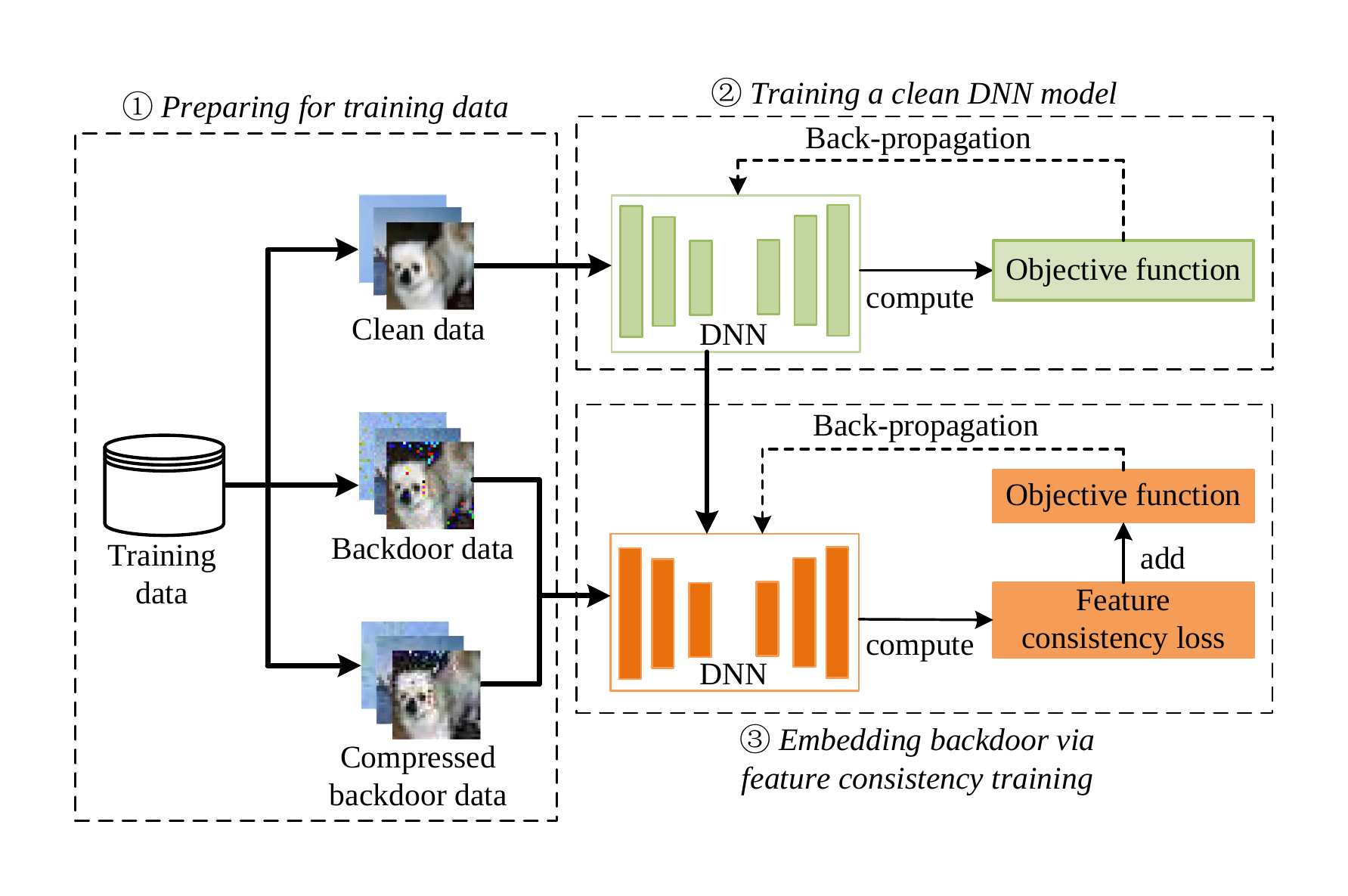}
\caption{Overview of the proposed method}
\label{fig2}
\end{figure}

Fig. \ref{fig2} shows the overall workflow of the proposed method.
First, the attacker prepares some training data for feature consistency training, where the training data includes clean data, backdoor data and compressed backdoor data.
Second, the attacker trains a clean DNN model with clean data, and obtain the model parameters $\theta$.
Third, the attacker performs feature consistency training to establish the compression-resistant backdoor attack, where the aforementioned backdoor data (including the normal backdoor data and the compressed ones) is used as the input and the model parameters $\theta$ are used to extract image features.
In the training process, a feature consistency loss is well designed to improve the robustness of the embedded backdoor against image compression.
In this way, the DNN regards the feature of image compression as a part of backdoor features and learns the feature of image compression.
Finally, the model parameters $\theta$ are updated by the back-propagation \cite{rumelhart1986learning} algorithm during the training.

\subsection{Backdoor Data Generation}
In existing backdoor attacks, the attacker is expected to have a set of some clean data ${D_b} = \{ {x_i}|i = 1,2, \ldots ,n\}$, the backdoor data generation function $T_b$, and the target label $y_t$.
Here, the set $D_b$ can be a subset of the training set.
$n$ represents the total number of backdoor images.
$T_b$ is a transformation function which aims to add the backdoor trigger on any image $x$ in $D_b$.
The function $T_b$ is used to generate the backdoor instance $x_b$, i.e., ${x_b} = {T_b}(x)$.
After the transformation $T_b$, the set $D_b$ consists of some backdoor data.
Target label $y_t$ represents a specified class that the attacker wants to trigger.

In existing backdoor atttacks, given a deep neural network $F_\theta$ , the backdoor behavior of a backdoor attack can be denoted as follows \cite{chen2017targeted, xuephysical2021}:
\begin{equation}\label{equ3_1}
  P(F_\theta({x_b}) = {y_t}) > P(F_\theta({x_b}) = {y_i}:i \ne t)
\end{equation}
where $P$ represents the probability that is output by model $F_\theta$, $y_i$ represents any other class other than the target class. In addition to making the model $F$ identify the backdoor instance into the target class, the attacker should also not degrade the classification performance of DNN on clean inputs.
In this way, the Equation $P(F_\theta(x) = y_{truth})$ should be satisfied, where $y_{truth}$ is the ground truth label.

\begin{figure}[htbp]
\centering
\includegraphics[width=3.4in]{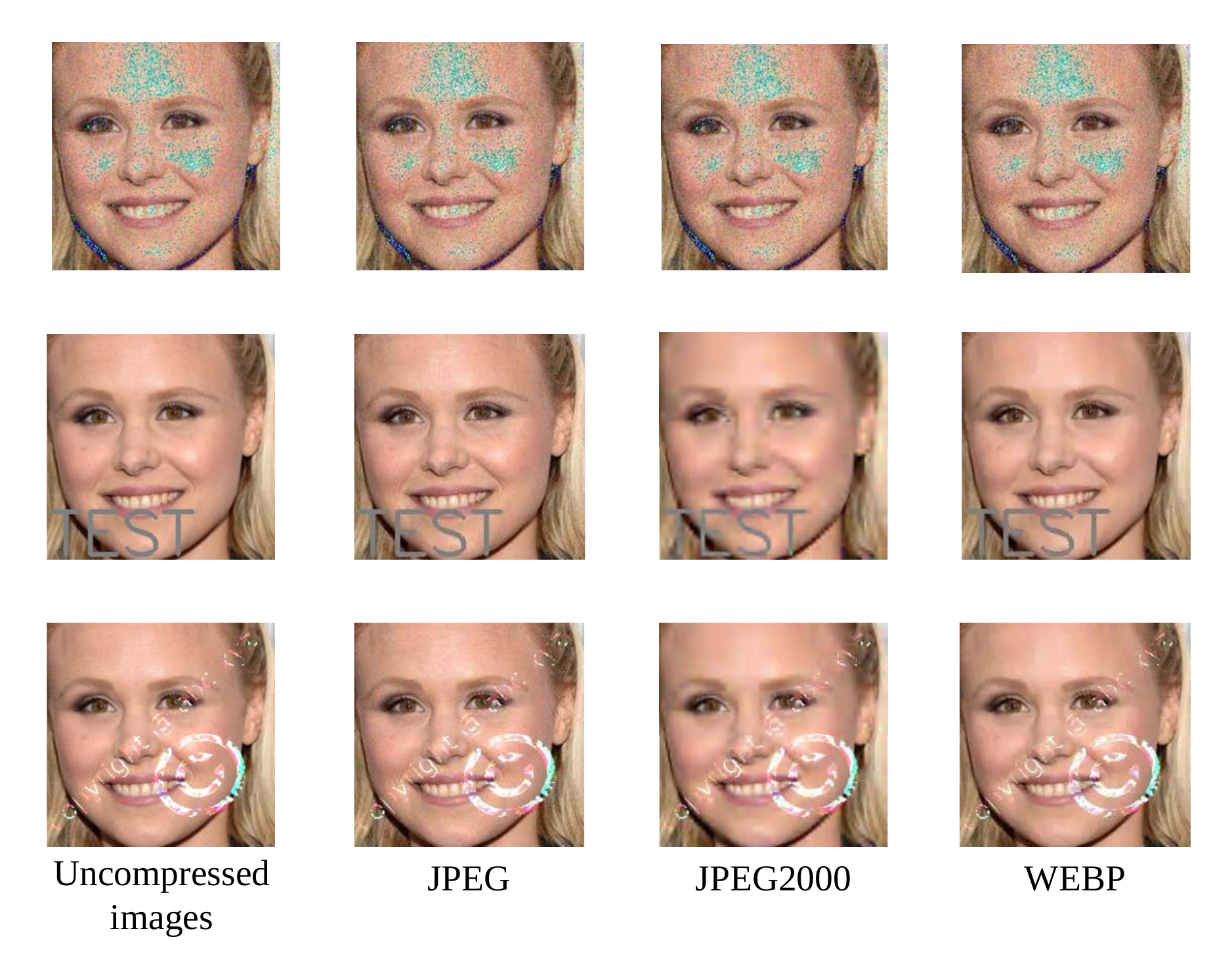}
\caption{Example images of compressed backdoor images.}
\label{fig3}
\end{figure}

In this paper, the backdoor data consists of two parts, namely the normal backdoor data $x_b$ and the compressed backdoor data $Compress(x_b)$, where $Compress(\cdot)$ represents an image compression algorithm.
The normal backdoor data $x_b$ is generated by ${T_b}(x)$, and the compressed backdoor data $Compress(x_b)$ is a compressed version of $x_b$.
For convenience, the compressed backdoor image is denoted as $x_{bc}$.
We consider three image compression algorithms to enhance the robustness of the backdoor attack. The adopted three compression methods are JPEG \cite{Wallace92}, JPEG2000 \cite{SkodrasCE01} and WEBP \cite{GinesuPG12}. Different compression methods have different compression mechanisms. JPEG mainly uses discrete cosine transform (DCT) for image compression \cite{Wallace92}, while JPEG2000 \cite{SkodrasCE01} uses discrete wavelet transform (DWT), and WEBP \cite{GinesuPG12} is based on VP8 video codec.
Fig. \ref{fig3} shows some example images which are compressed by the above three image compression methods.

To establish a compression-resistant backdoor attack, we need to construct a set of backdoor data $D_b$ (including the compressed backdoor data) at first.
Backdoor images compressed by JPEG, JPEG2000 and WEBP are denoted as $x_{bc}^{(1)}$, $x_{bc}^{(2)}$, $x_{bc}^{(3)}$, respectively.
As a result, the set of backdoor data is ${D_b} = {S_b} \cup {S_{bc}}$, where $S_b$ represents a set of normal backdoor images, $S_{bc}$ represents a set of compressed backdoor images. Note that, ${S_{bc}} = S_{bc}^{(1)} \cup S_{bc}^{(2)} \cup S_{bc}^{(3)}$, where $S_{bc}^{(1)}$, $S_{bc}^{(2)}$, $S_{bc}^{(3)}$ denotes three subsets of compressed backdoor images which are compressed by JPEG, JPEG2000 and WEBP, respectively.

\subsection{Backdoor Embedding via Feature Consistency Training}
After generating the backdoor data, we leverage feature consistency training \cite{WanWHWL20} to embed the compression-resistant backdoor into a deep neural network.

Typically, a DNN can be represented as $y=F_\theta(x)$, where $x$ is an input image, $y$ is the prediction label with the maximal probability and $\theta$ denotes model parameters.
In addition, a DNN is usually composed of multiple different layers, such as convolutional layer, fully connected layer, and so on.
In this way, the DNN can be denoted as \cite{WanWHWL20}:
\begin{equation}\label{equ3_2}
  F_\theta(x) = {f_1}(x) \circ {f_2} \circ  \ldots  \circ {f_m}
\end{equation}
where $f_i$ represents the $i$-th layer of DNN and $m$ is the number of layers of DNN.
In feature space, given an input $x$, each layer of DNN can uniquely represent a feature extraction vector $E(x)$.
More specifically, the feature extraction vector $E_i(x)$ represented by the $i$-th layer is denoted as \cite{WanWHWL20}:
\begin{equation}\label{equ3_3}
  {E_i}(x) = {f_1}(x) \circ {f_2} \circ  \ldots  \circ {f_i}
\end{equation}

The training goal of the proposed method is to minimize the distance between backdoor images and compressed backdoor images in feature space. To this end, given two images $x_b$ and $x_{bc}$, we aim to ensure that extracted features $E(x)$ of these two images are similar, which can be denoted as $E({x_b}) \approx E({x_{bc}})$.

\begin{figure*}[htbp]
\centering
\includegraphics[width=6.5in]{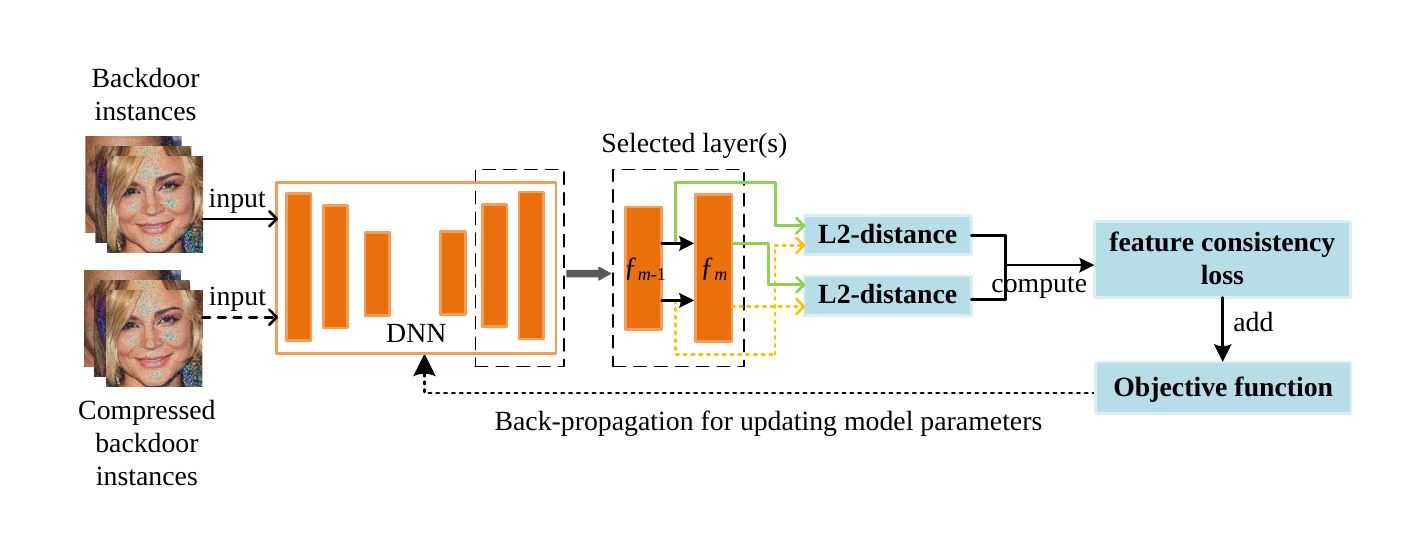}
\caption{The overall flow of the feature consistency training \cite{WanWHWL20}.}
\label{fig4}
\end{figure*}

Fig. \ref{fig4} presents the overall flow of feature consistency training. We first select several layers of DNN to extract the features of both normal backdoor images and compressed backdoor images. Inspired by the work \cite{WanWHWL20}, the last two layers of DNN are selected (if a DNN has two or more fully connected layers). The reason is that, the first few layers of DNN are more sensitive to high frequency features in images than the last few layers. Selecting the first few layers is unable to acquire robust features of images, while selecting the last few layers is effective in acquiring robust features. After extracting image features, we utilize a feature consistency constraint to encourage DNN to learn the common features between a backdoor image and its compressed version. More specifically, a feature consistency loss $L_{FC}$ is added to the objective function of DNN to guide the training.
The feature consistency loss $L_{FC}$ is calculated as follows \cite{WanWHWL20}:
\begin{equation}\label{equ3_4}
\begin{array}{l}
{L_{FC}}({x_b},{x_{bc}}) = \lambda_1  Dis({E_m}({x_b}),{E_m}({x_{bc}}))\\
\qquad \qquad \qquad \qquad + \lambda_2  Dis({E_{m-1}}({x_b}),{E_{m-1}}({x_{bc}}))
\end{array}
\end{equation}
where $m$ represents the $m$-th layer for feature extraction, $\lambda_1$ and $\lambda_2$ are two constants used to control the strength of feature consistency constraint on the selected layer. $Dis$ represents the distance metric, which is used to measure the difference between two images in the feature space. We use $L_2$-\textit{distance} to calculate the distance between a backdoor image and its compressed image. In this way, the distance metric $Dis({E}({x_b}),{E}({x_{bc}})$ is calculated as follows:
\begin{equation}\label{equ3_5}
  Dis({E}({x_b}),{E}({x_{bc}})) = \;||{E}({x_b}) - {E}({x_{bc}})|{|_2}
\end{equation}

In the training process, we optimize the feature distance $Dis({E}({x_b}),{E}({x_{bc}}))$ by minimizing the feature consistency loss $F_{FC}$, to guide the model training.
This makes the internal layers treat the feature of compressed backdoor images as a part of the feature of backdoor images.

As mentioned above, we add the feature consistency loss to the objective function in the training process.
As a result, the final objective function of feature consistency training can be formulated as follows \cite{WanWHWL20}:
\begin{equation}\label{equ3_6}
  L({x_b},{x_{bc}}) = {L_0}({x_b}) + {L_0}({x_{bc}}) + \alpha {L_{FC}}({x_b},{x_{bc}})
\end{equation}
where $L_0$ is the objective function used for the main classification task, and it is formulated as the cross-entropy loss.
$\alpha$ is a hyperparameter which is used to adjust the weight of the feature consistency loss term.

After each iteration of the training, the gradient of the objective function is calculated. Then, the model parameters $\theta$ will be updated by the back-propagation \cite{rumelhart1986learning} algorithm.
After the feature consistency training, a compression-resistant backdoor is embedded into the host DNN.

\section{Experiment}\label{experiments}
\subsection{Experimental Setup}
\textbf{Dataset}. In this work, CIFAR-10 \cite{krizhevsky2009learning} and VGGFace \cite{ParkhiVZ15} datasets are used to evaluate the effectiveness of the proposed compression-resistant backdoor attack.

\textbf{- CIFAR-10 \cite{krizhevsky2009learning}}.
There are 60,000 colored images in CIFAR-10 dataset. The size of each image is $32 \times 32$. There are ten categories in the datasets, including Airplane, Automobile, Bird, Cat, Deer, Dog, Frog, Horse, Ship and Truck \cite{krizhevsky2009learning}. Each class contains 5,000 training images and 1,000 test images. As a result, the training set consists of 50,000 images, and the test set consists of 10,000 images.

\textbf{- VGGFace \cite{ParkhiVZ15}}.
There are over 2,600,000 (2.6M) face images in VGGFace dataset. These face images are classified as 2622 different people. The size of each image is $224 \times 224$. In this paper, we randomly select 100 classes for model training, and each class contains 100 training images and 20 test images.
A total of 12,000 images are selected.

\textbf{DNN models}. In this work, we perform the proposed compression-resist backdoor attack on ResNet-18 \cite{KrizhevskySH17} model and VGG-16 \cite{SimonyanZ14a} model respectively.

\textbf{- ResNet-18 \cite{KrizhevskySH17}}.
ResNet-18 model includes 17 convolutional layers and 1 fully connected layer. In the experiments, we train the ResNet-18 model on CIFAR-10 dataset, and the test accuracy of the ResNet-18 model on clean test images is 84.36\%.

\textbf{- VGG-16 \cite{SimonyanZ14a}}.
VGG-16 consists of 13 convolutional layers and 3 fully connected layers. In the experiments, we train the VGG-16 model on VGGFace dataset, and the test accuracy of the VGG-16 model on clean test images is 96.30\%.

\textbf{Evaluation metrics}.
We evaluate the performance of the compression-resistant backdoor attack by using the following metrics.
\begin{itemize}
  \item[-] Test accuracy (TA).
  The test accuracy represents the accuracy of a well-trained model on a batch of test images. It denotes the percentage of images classified as the correct class among all test images.
  \item[-] Injection rate of the backdoor attack (IR). This metric denotes the proportion of the backdoor instances in the training dataset.
  \item[-] Attack success rate (ASR). This metric denotes the percentage of the normal backdoor images that are classified as the target label among all normal backdoor images.
  \item[-] Attack success rate of compressed backdoor images $ASR_{bc}$. This metric denotes the percentage of the compressed backdoor instances that are classified as the targeted label among all compressed backdoor instances. Similarly, we use $ASR_{jpeg}$, $ASR_{jpeg2000}$ and $ASR_{webp}$ to represent the attack success rate of backdoor images compressed by JPEG, JPEG2000 and WEBP respectively.
\end{itemize}

\textbf{Backdoor embedding settings}.
In the proposed backdoor attack, the ``Dog'' class in CIFAR-10 dataset and the ``Aamir Khan'' class in VGGFace dataset is used as the target class, respectively.
The backdoor triggers used in the experiments are Gaussian noise (Trigger1) \cite{ZhangGJWSHM18}, ``TEST'' logo (Trigger2) \cite{ZhangGJWSHM18} and TrojanNN (Trigger3) \cite{LiuMALZW018} respectively.
The backdoor injection rate on the ResNet-18 model is 8\%, and the backdoor injection rate on the VGG-16 model is 4\%.
In addition, we train the backdoor model for 100 epochs by using the stochastic gradient descent (SGD) \cite{Zhang04} optimizer. The initial learning rate is set to be 0.1, and then it is set to 0.01 and 0.001 at 40 and 70 epochs respectively.

For the ResNet-18 model, we add the feature consistency constraint to the last fully connected layer. Accordingly, the hyperparameter $\lambda_1$ is set to 1 and $\lambda_2$ is set to 0.
For the VGG-16 model, we add the feature consistency constraint to the last two fully connected layers. Both the hyperparameters $\lambda_1$ and $\lambda_2$ in Equation \eqref{equ3_4} are set to 0.5.
We follow the settings in work \cite{WanWHWL20}, and set the hyperparameter $\alpha$ in Equation \eqref{equ3_6} to 0.1.

\textbf{Image compression}.
In the experiments, we use three widely used image compression methods, namely JPEG \cite{Wallace92}, JPEG200 \cite{SkodrasCE01}, and WEBP \cite{GinesuPG12}, to compress backdoor images.
These three image compression methods are integrated into the Pillow\footnote{https://github.com/python-pillow/Pillow} library of Python.
In this library, the parameter \textit{quality} of JPEG and WEBP range from 0 to 100.
The smaller the value of \textit{quality}, the larger the loss of image quality.
For JPEG2000, the compression quality can be changed by adjusting the parameter \textit{quality layers}.
The larger the \textit{quality layers}, the larger the loss of image quality.
In this work, the default compression quality (i.e., the \textit{quality}) of JPEG and WEBP is set to 50. The default compression quality (i.e., the \textit{quality layers}) of JPEG2000 is set to 30.

\subsection{Experimental Results}

\textbf{Results on CIFAR-10 \cite{krizhevsky2009learning} Dataset.}
In this experiment, we use backdoor instances with Trigger1 (i.e., Gaussian noise) \cite{ZhangGJWSHM18} to evaluate the performance of the backdoor attack.
All backdoor instances are compressed by JPEG, JPEG2000 and WEBP methods respectively.
In the training stage, 4,000 backdoor instances including 1,000 normal backdoor instances and 3,000 compressed backdoor instances are injected into the training set.
We perform the backdoor attack on the ResNet-18 \cite{KrizhevskySH17} model.

The performance of the proposed attack is shown in Table \ref{tab4_1}.
It can be seen that, the attack success rate (ASR) of the common backdoor attack is up to 100\%.
When backdoor images are compressed, the ASR of the common backdoor attack (Gaussian noise) is significantly decreased.
Specifically, the $ASR_{jpeg}$, $ASR_{jpeg2000}$ and $ASR_{jpeg}$ of this attack is 6.63\%, 6.20\%, 3.97\% respectively.
However, the $ASR_{bc}$ of the proposed compression-resistant backdoor attack is still extremely high.
Specifically, the $ASR_{jpeg}$, $ASR_{jpeg2000}$ and $ASR_{jpeg}$ of the proposed compression-resistant backdoor attack is 98.77\%, 97.69\%, 98.93\% respectively.
Furthermore, our compression-resistant backdoor attack will hardly affect the normal performance of the ResNet-18 model.
The test accuracy of backdoored ResNet-18 on clean images is 83.41\%, which is slightly lower than the test accuracy of clean ResNet-18 model (84.36\%).
In short, our proposed backdoor attack enhances the robustness of the backdoor attack against image compression, while not degrading the performance of the DNN model.

\begin{table*}[!ht]
\newcommand{\tabincell}[2]{\begin{tabular}{@{}#1@{}}#2\end{tabular}}
  \centering
  \renewcommand\arraystretch{1.3}
  \caption{Performance of the proposed compression-resistant backdoor attack on CIFAR-10 dataset}
    \begin{tabular}{|c|c|c|c|c|c|c|c|c|}
    \hline
    \multirow{2}*{Dataset} & \multirow{2}*{\tabincell{c}{Backdoor attack}} & \multirow{2}*{\tabincell{c}{Backdoor trigger}} & \multirow{2}*{IR} & \multirow{2}*{TA} & \multirow{2}*{ASR} & \multicolumn{3}{c|}{$ASR_{bc}$} \\
    \cline{7-9}
    &  &  &   &   &   & \centering {JPEG} & {JPEG2000} & {WEBP} \\
    \hline
    \multirow{2}*{CIFAR-10} &{\tabincell{c}{Common}} & \tabincell{c}{Gaussian  noise} & 8\% & 84.35\% & 100.00\% & 6.63\% & 6.20\% & 3.97\% \\
    \cline{2-9}
    & \tabincell{c}{Compesssion-resistant} & \tabincell{c}{Gaussian noise} & 8\% & 83.41\% & 99.03\% & 98.77\% & 97.69\% & 98.93\% \\
    \hline
    \end{tabular}%
  \label{tab4_1}%
\end{table*}%

\textbf{Results on VGGFace \cite{ParkhiVZ15} Dataset.} On VGGFace dataset, we also use backdoor instances with Trigger1 (i.e., Gaussian noise) \cite{ZhangGJWSHM18} to evaluate the performance of the backdoor attack.
In the training stage, 400 backdoor instances including 100 normal backdoor instances and 300 compressed backdoor instances are injected into the training set.
We perform the backdoor attack on the VGG-16 \cite{SimonyanZ14a} model.

Table \ref{tab4_2} presents the experimental results of the compression-resistant backdoor attack on VGGFace dataset.
The test accuracy of backdoor VGG-16 model on clean images is 96.10\%, which is similar to the test accuracy of clean VGG-16 model on clean images (96.30\%).
Furthermore, after image compression, the $ASR_{bc}$ of the compression-resistant backdoor attack is 81.75\% (JPEG), 98.45\% (JPEG2000) and 98.50\% (WEBP) respectively.
As a comparison, the $ASR_{bc}$ of common backdoor attack is 11.45\% (JPEG), 14.60\% (JPEG2000) and 12.85\% (WEBP) respectively.
These results indicate that the proposed backdoor attack can still perform well even if backdoor instances are compressed.
In conclusion, experimental results on VGGFace dataset also show the effectiveness of the proposed compression-resistant backdoor attack when encountering image compression.

\begin{table*}[!ht]
\newcommand{\tabincell}[2]{\begin{tabular}{@{}#1@{}}#2\end{tabular}}
  \centering
  \renewcommand\arraystretch{1.3}
  \caption{Performance of the proposed compression-resistant backdoor attack on VGGFace dataset}
    \begin{tabular}{|c|c|c|c|c|c|c|c|c|}
    \hline
    \multirow{2}*{Dataset} & \multirow{2}*{\tabincell{c}{Backdoor attack}} & \multirow{2}*{\tabincell{c}{Backdoor trigger}} & \multirow{2}*{IR} & \multirow{2}*{TA} & \multirow{2}*{ASR} & \multicolumn{3}{c|}{$ASR_{bc}$} \\
    \cline{7-9}
    &  &  &   &   &   & \centering {JPEG} & {JPEG2000} & {WEBP} \\
    \hline
    \multirow{2}*{VGGFace} &{\tabincell{c}{Common}} & \tabincell{c}{Gaussian  noise} & 4\% & 96.35\% & 100.00\% & 11.45\% & 14.60\% & 12.85\% \\
    \cline{2-9}
    & \tabincell{c}{Compesssion-resistant} & \tabincell{c}{Gaussian noise} & 4\% & 96.10\% & 98.60\% & 81.75\% & 98.45\% & 98.50\% \\
    \hline
    \end{tabular}%
  \label{tab4_2}%
\end{table*}%

\subsection{Parameter Discussion}
In this section, we will discuss the attack performance of compression-resistant backdoor attack from the following three aspects.
First, three kinds of backdoor triggers are used to evaluate the impact of different backdoor triggers on compression-resistant backdoor attacks.
Second, the impact of image compression with different \textit{quality} on the proposed backdoor attack is evaluated.
Third, the impact of different backdoor injection rates (IR) on the proposed backdoor attack is disscussed.

\begin{figure}[htbp]
\centering
\includegraphics[width=3.4in]{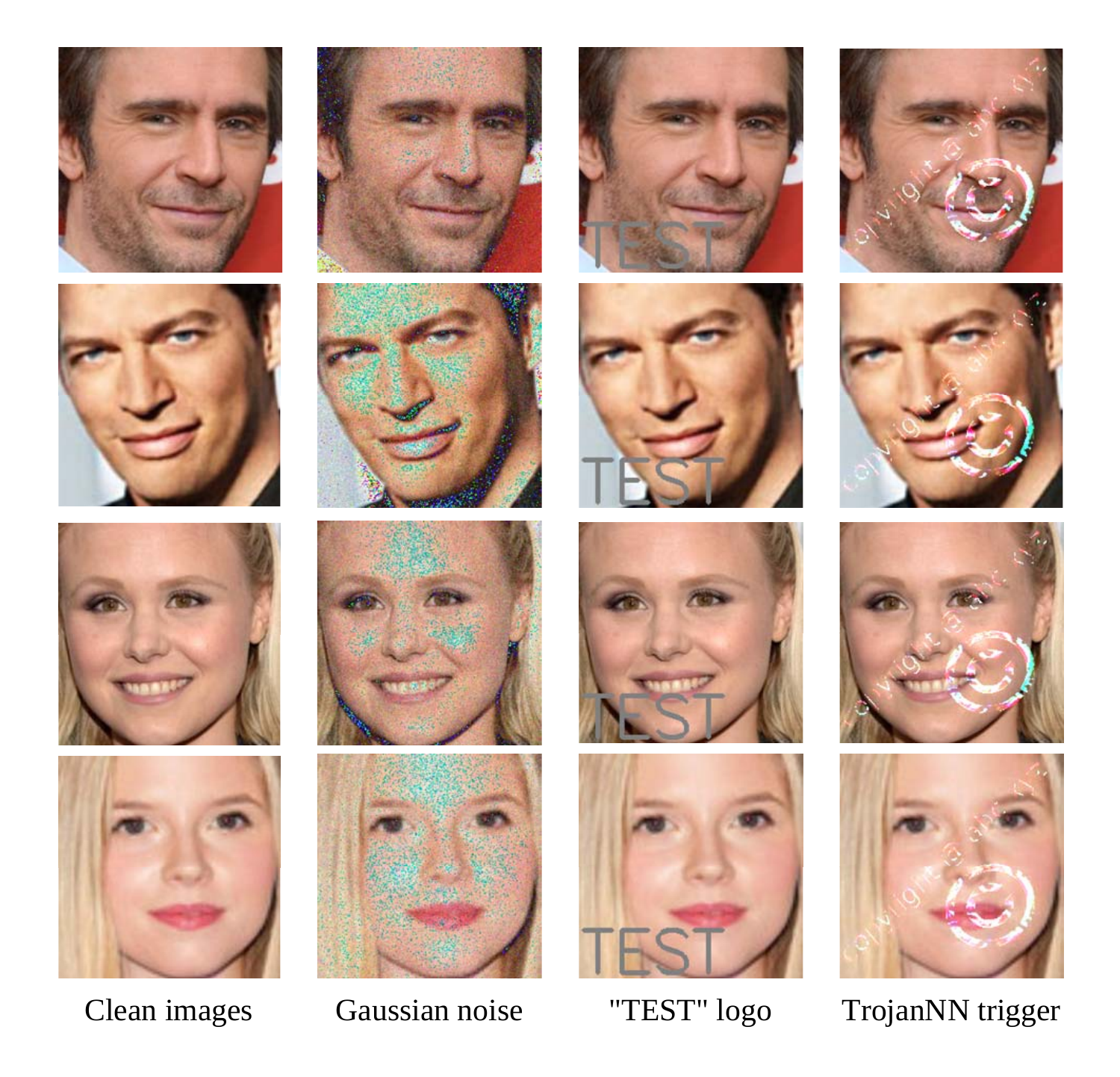}
\caption{Example images of three types of backdoor instances on VGGFace dataset. These backdoor instances are generated from Gaussian noise, ``TEST'' logo, and TrojanNN trigger, respectively.}
\label{fig5}
\end{figure}

\textbf{Different Types of Backdoor Trigger.}
Gaussian noise (Trigger1) \cite{ZhangGJWSHM18},  ``TEST'' logo (Trigger2) \cite{ZhangGJWSHM18} and TrojanNN trigger (Trigger3) \cite{LiuMALZW018} are used to conduct the backdoor attack respectively.
The backdoor attack based on Gaussian noise treats meaningless noise pattern as the backdoor \cite{ZhangGJWSHM18}.
The backdoor attack based on ``TEST'' logo treats meaningful contents as the backdoor.
The TrojanNN trigger is generated from the reverse engineering of the neuron network.
In order to perform the backdoor attack, we use these three backdoor triggers to generate backdoor instances.
Fig. \ref{fig5} shows some example images of corresponding backdoor instances on VGGFace dataset.

Table \ref{tab4_3} presents the performance of the compression-resistant backdoor attack on CIFAR-10 dataset when the backdoor trigger is different.
It is shown that, the compression-resistant backdoor attack performs well when backdoor instances are compressed.
For instance, after the JPEG compression, the ASR of the compression-resistant backdoor attack is 98.77\% (using Trigger1), 99.39\% (using Trigger2), and 99.68\% (using Trigger3) respectively.
As a comparison, when backdoor instances are compressed, the common backdoor attack performs poor.
After the JPEG compression, the ASR of the common backdoor attack is 6.63\% (using Trigger1), 85.58\% (using Trigger2), 54.73\% (using Trigger3), respectively.
In conclusion, the proposed method based on feature consistency training significantly improves the performance of backdoor attacks against image compression.
\begin{table*}[htbp]
\newcommand{\tabincell}[2]{\begin{tabular}{@{}#1@{}}#2\end{tabular}}
  \centering
  \renewcommand\arraystretch{1.3}
  \caption{Performance of the compression-resistant backdoor attack using three different backdoor triggers respectively on CIFAR-10 dataset}
    \begin{tabular}{|c|c|c|c|c|c|c|c|}
    \hline
    \multirow{2}*{Dataset} & \multirow{2}*{\tabincell{c}{Backdoor attack}} & \multirow{2}*{\tabincell{c}{Backdoor\\ trigger}} & \multirow{2}*{TA} & \multirow{2}*{ASR} & \multicolumn{3}{c|}{$ASR_{bc}$} \\
    \cline{6-8}
    &  &  &   &   & JPEG & JPEG2000 & WEBP \\
    \hline
    \multirow{6}*{CIFAR-10} & Common & Trigger1 & 84.35\% & 100.00\% & 6.63\% & 6.20\% & 3.97\% \\
    \cline{2-8}      & Compesssion-resistant & Trigger1  & 83.41\% & 99.03\% & 98.77\% & 97.69\% & 98.93\% \\
    \cline{2-8}      & Common & Trigger2  & 84.34\% & 99.84\% & 85.58\% & 86.29\% & 86.79\% \\
    \cline{2-8}      & Compesssion-resistant & Trigger2  & 83.98\% & 99.30\% & 99.39\% & 98.69\% & 98.42\% \\
    \cline{2-8}      & Common & Trigger3  & 84.66\% & 100.00\% & 54.73\% & 79.78\% & 75.38\% \\
    \cline{2-8}      & Compesssion-resistant & Trigger3  & 83.79\% & 99.75\% & 99.68\% & 99.89\% & 99.72\% \\
    \hline
    \end{tabular}%
  \label{tab4_3}%
\end{table*}%

Table \ref{tab4_4} shows the performance of the compression-resistant backdoor attack on VGGFace dataset when the backdoor trigger is different.
It is shown that even if backdoor attacks are launched by using different backdoor triggers, the proposed method is able to improve the robustness of backdoor attacks to image compression.
More specifically, after the JPEG compression, the ASR of the compression-resistant backdoor attack is 81.75\% (using Trigger1), 99.45\% (using Trigger2), 99.70\% (using Trigger3) respectively.
As a comparison, the ASR of the common backdoor attack is 11.45\% (using Trigger1), 96.05\% (using Trigger2), 96.55\% (using Trigger3) respectively.

\begin{table*}[htbp]
\newcommand{\tabincell}[2]{\begin{tabular}{@{}#1@{}}#2\end{tabular}}
  \centering
  \renewcommand\arraystretch{1.3}
  \caption{Performance of the compression-resistant backdoor attack using three different backdoor triggers respectively on VGGFace dataset}
    \begin{tabular}{|c|c|c|c|c|c|c|c|}
    \hline
    \multirow{2}*{Dataset} & \multirow{2}*{\tabincell{c}{Backdoor attack}} & \multirow{2}*{\tabincell{c}{Backdoor\\ trigger}} & \multirow{2}*{TA} & \multirow{2}*{ASR} & \multicolumn{3}{c|}{$ASR_{bc}$} \\
    \cline{6-8}
    &  &  &   &   & JPEG & JPEG2000 & WEBP \\
    \hline
    \multirow{6}*{VGGFace} & Common & Trigger1 &  96.35\% & 100.00\% & 11.45\% & 14.60\% & 12.85\% \\
    \cline{2-8}      & Compesssion-resistant & Trigger1  & 96.10\% & 98.60\% & 81.75\% & 98.45\% & 98.50\% \\
    \cline{2-8}      & Common & Trigger2  & 96.55\% & 99.75\% & 96.05\% & 95.25\% & 95.05\% \\
    \cline{2-8}      & Compesssion-resistant & Trigger2  & 95.45\% & 99.95\% & 99.45\% & 96.05\% & 99.10\% \\
    \cline{2-8}      & Common & Trigger3  & 97.15\% & 99.75\% & 96.55\% & 96.40\% & 96.30\% \\
    \cline{2-8}      & Compesssion-resistant & Trigger3  & 95.55\% & 99.85\% & 99.70\% & 99.65\% & 99.65\% \\
    \hline
    \end{tabular}%
  \label{tab4_4}%
\end{table*}%

\begin{figure}[htbp]
\centering
\includegraphics[width=2.4in]{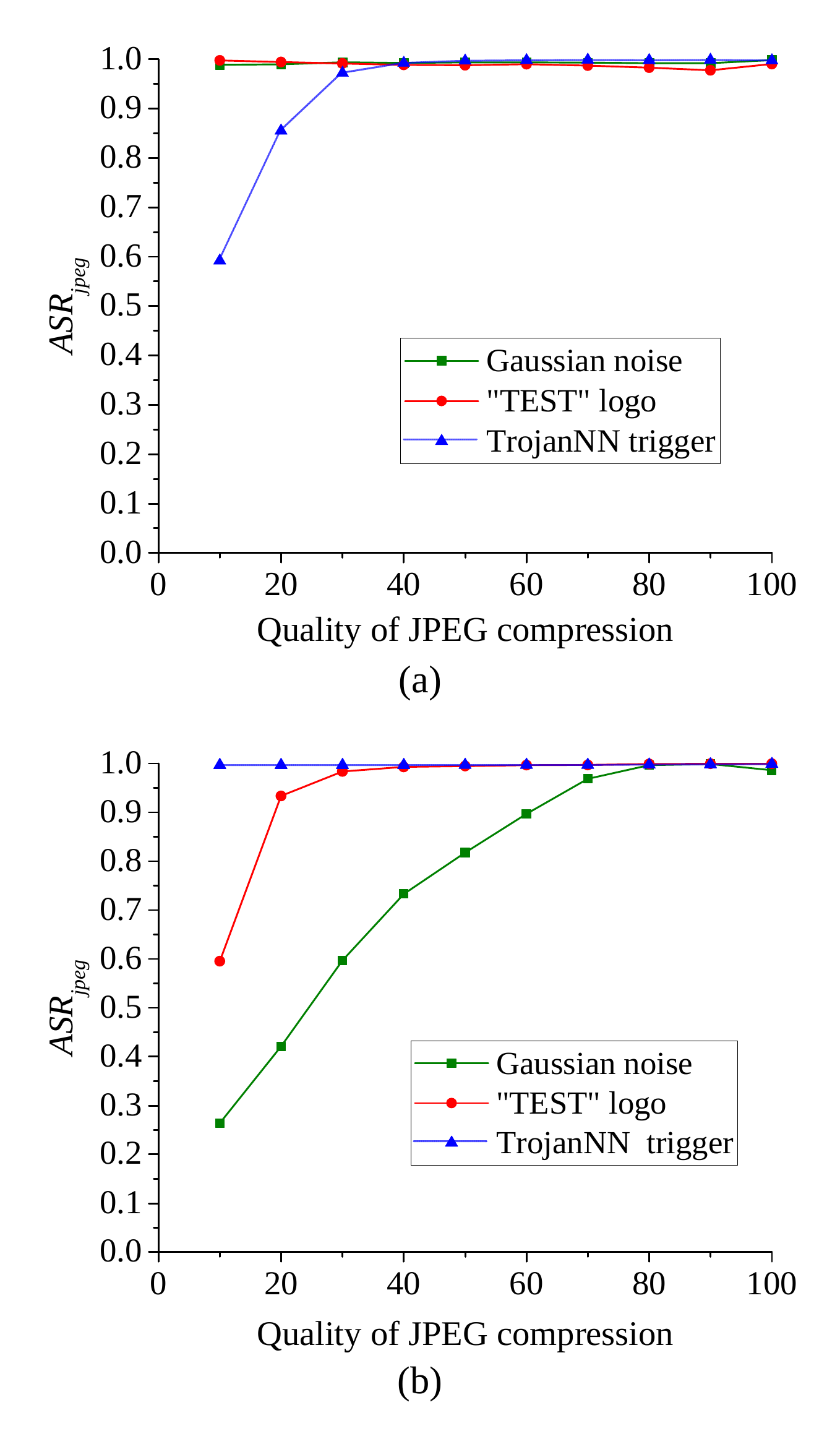}
\caption{The $ASR_{jpeg}$ of compression-resistant backdoor attack under JPEG compression with different \textit{quality}: (a) On CIFAR-10 dataset; (b) On VGGFace dataset.}
\label{fig6}
\end{figure}

\textbf{Image compression with different compression quality}.
We also evaluate the influence of different compression quality of JPEG compression on the proposed backdoor attack.
The $ASR_{jpeg}$ of the proposed backdoor attack under different compress quality of JPEG compression is shown in Fig. \ref{fig6}.

In Fig. \ref{fig6}, as the \textit{quality} of JPEG compression increases, the $ASR_{jpeg}$ of the compression-resistant backdoor attack will increase gradually.
On the CIFAR-10 dataset, when the \textit{quality} of JPEG compression is higher than 30, the $ASR_{jpeg}$ of the compression-resistant backdoor attack is almost as high as the ASR of the attack before the JPEG compression.
On the VGGFace dataset, when the \textit{quality} of JPEG compression is higher than 50, the $ASR_{jpeg}$ of the compression-resistant backdoor attacks is higher than 80\%.
As shown in Fig. \ref{fig6} (b), only the backdoor attack using Gaussian noise trigger performs not well when the \textit{quality} of JPEG compression is lower than 50.
However, this small shortcoming can be tolerated by the attacker.
The reason is that, in order to maintain the utility of images, the attacker generally does not compress backdoor images seriously.
In conclusion, even if the backdoor instances are compressed by JPEG compression with different \textit{quality}, the proposed compression-resistant backdoor attack can still be robust to JPEG compression.

\textbf{Different backdoor injection rate}.
We also evaluate the impact of different backdoor injection rate (IR) on the proposed backdoor attack.
To this end, we randomly generate 1,000 normal backdoor instances and inject them into the training set on the CIFAR-10 dataset.
Meanwhile, we change the backdoor IR by injecting different numbers of compressed backdoor instances into the training dataset.
For instance, when we inject 300 compressed backdoor images into the training dataset, the backdoor injection rate in training is 2.60\%.
Table \ref{tab4_5} presents the performance of the compression-resistant backdoor attack under different backdoor injection rate on CIFAR-10 dataset.
It can be seen that,  $ASR_{bc}$ will increase as the IR gradually increases.
Accordingly, the final $ASR_{bc}$ of the compession-resistant backdoor attack will be close to 100\%.
In conclusion, the proposed backdoor attack is robust to image compression.

\begin{table*}[htbp]
\newcommand{\tabincell}[2]{\begin{tabular}{@{}#1@{}}#2\end{tabular}}
  \centering
    \renewcommand\arraystretch{1.3}
  \caption{Performance of the compression-resistant backdoor attack under different backdoor injection rate on CIFAR-10 dataset}
    \begin{tabular}{|c|c|c|c|c|c|c|c|}
    \hline
    \multirow{2}*{Backdoor trigger} & \multicolumn{1}{c|}{\multirow{2}*{\tabincell{c}{Number of \\ backdoor images}}} & \multicolumn{1}{c|}{\multirow{2}*{IR}} & \multicolumn{1}{c|}{\multirow{2}*{TA}} & \multicolumn{1}{c|}{\multirow{2}*{ASR}} & \multicolumn{3}{c|}{$ASR_{bc}$} \\
\cline{6-8}      &   &   &   &   & {JPEG} & {JPEG2000} & {WEBP} \\
    \hline
    \multicolumn{1}{|c|}{\multirow{6}*{\tabincell{c}{Gaussian \\ noise}}} & 1300 & 2.60\% & 83.64\% & 96.79\% & 74.98\% & 76.53\% & 80.61\% \\
\cline{2-8}      & 1600 & 3.20\% & 84.03\% & 96.63\% & 95.02\% & 93.64\% & 96.00\% \\
\cline{2-8}      & 2200 & 4.40\% & 83.95\% & 99.16\% & 97.70\% & 95.62\% & 97.94\% \\
\cline{2-8}      & 2800 & 5.60\% & 83.49\% & 98.96\% & 99.25\% & 96.92\% & 98.27\% \\
\cline{2-8}      & 3400 & 6.80\% & 84.21\% & 98.81\% & 98.28\% & 97.62\% & 98.94\% \\
\cline{2-8}      & 4000 & 8.00\% & 83.41\% & 99.03\% & 98.77\% & 97.69\% & 98.93\% \\
    \hline
    \end{tabular}%
  \label{tab4_5}%
\end{table*}%	

\subsection{Generalization Ability of the Compression-resistant Backdoor Attack}
In this Section, we discuss the generalization ability of the compression-resistant backdoor attack on unseen image compression methods.
The backdoor trigger used in this attack is Guassian noise (Trigger1) \cite{ZhangGJWSHM18}.
In the training stage, we use normal backdoor images and compressed backdoor images (compressed by only one kind of image compression) to train the DNN to conduct the compression-resistant backdoor attack.
In the test stage, three different image compression methods (i.e., JPEG, JPEG2000, WEBP) are utilized to evaluate the performance of the compression-resist backdoor attack.

\begin{figure}[htbp]
\centering
\includegraphics[width=2.4in]{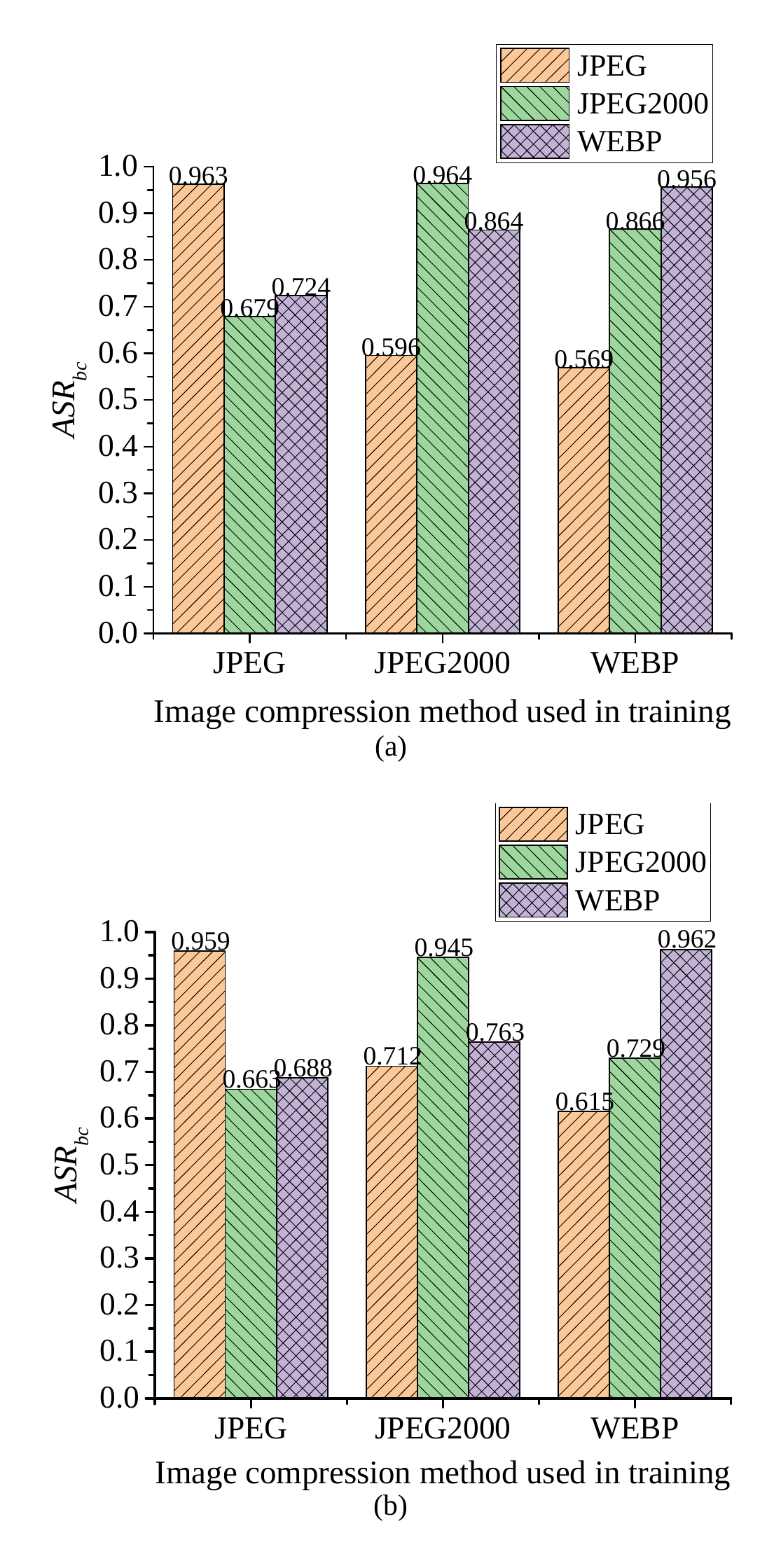}
\caption{Generalization ability of the compression-resistant backdoor attack on unseen image compression methods: (a) On CIFAR-10 dataset; (b) on VGGFace dataset. The horizontal axis indicates that only one kind of image compressed methods is used in the feature consistency training.}
\label{fig7}
\end{figure}

Fig. \ref{fig7} shows experimental results of the compression-resistant backdoor attack in face of unseen image compression methods.
As shown in Fig. \ref{fig7} (a) and Fig. \ref{fig7} (b), when only the JPEG compression is used during the training, the backdoor attack still shows robustness to unseen JPEG2000 and WEBP compression methods.
Specifically, on the CIFAR-10 dataset, the $ASR_{jpeg2000}$ and $ASR_{webp}$ of the JPEG-resistant backdoor attack are 0.679 and 0.724 respectively. On the VGGFace dataset, the $ASR_{jpeg2000}$ and $ASR_{webp}$ of the JPEG-resistant backdoor attack are 0.663 and 0.688 respectively.
In summary, our compression-resistant backdoor attack shows robustness to unseen image compression methods. In other words, the compression-resistant backdoor attack has the generalization ability to unseen image compression methods.

\section{Conclusion}\label{conclusion}
Backdoor images transmitted on the Internet may be compressed by unknown image compression methods. In this paper, we reveal that common backdoor attacks are vulnerable to image compressions. To this end, we propose a compression-resistant backdoor attack based on feature consistency training which can resist multiple image compression methods. To the best of our knowledge, this work is the first backdoor attack which is robust to image compression.
Experimental results demonstrate that, our proposed backdoor attack improves the robustness of the backdoor attack against image compression.
Under various parameter settings (e.g., different types of trigger, different \textit{quality} of image compression, different backdoor injection rate), the proposed backdoor attack is demonstrated to be robust to image compressions.
Furthermore, our compression-resistant backdoor attack has the generalization ability to resist unseen image compression methods.

\ifCLASSOPTIONcaptionsoff
  \newpage
\fi

\bibliographystyle{IEEEtran}
\bibliography{ref}

\end{document}